# GPT as ghostwriter at the White House

**Compared with four US presidents, what are the differences between *State of the Union* addresses written by GPT and the true presidents**

Jacques Savoy

Computer Science Dept.
University of Neuchatel
rue Emile Argand 11
2000 Neuchatel, Switzerland
Jacques.Savoy@unine.ch

**Abstract**

Recently several large language models (LLMs) have demonstrated their capability to generate a message in response to a user request. Such scientific breakthroughs promote new perspectives but also some fears. The main focus of this study is to analyze the written style of one LLM called ChatGPT 3.5 by comparing its generated messages with those of the recent US presidents. To achieve this objective, we compare the *State of the Union* addresses written by Reagan to Obama with those automatically produced by ChatGPT. We found that ChatGPT tends to overuse the lemma "we" as well as nouns and commas. On the other hand, the generated speeches employ less verbs and include, in mean, longer sentences. Even when imposing a given style to ChatGPT, the resulting speech remains distinct from messages written by the target author. Moreover, ChatGPT opts for a neutral tone with mainly positive emotional expressions and symbolic terms (e.g., freedom, nation). Finally, we show that the GPT's style exposes distinct features compared to real presidential addresses.

## 1   Introduction

Based on large language models (LLMs) (Zhoa *et al*., 2023), the computer is able to provide an explanation to users' queries. These LLMs (e.g., OpenAI's GPT, Meta's Llama, Google's Bard, Gemini, PaLM or T5) can maintain a dialogue, for example, to help the user in identifying or resolving a problem. To generate such text, LLMs have been trained with massive web-text data, newspapers[1] and books corpora. Of course, the value of the generated message depends on the quantity and quality of those sources as well as various intern choices (neural network architecture, parameters, optimization algorithm, etc.).

Even if authoring short texts corresponds to the primary LLM application, the solution of other problems have been put forward such as automatic translation (Jiao *et al*., 2023), solving mathematical problems, producing music, coding simple applications, generating images in a given script language, etc. (Bubeck *et al*., 2023). However, the generated output may encompass hallucinations, a subject not covered in the current study. From a stylistic point of view, Bubeck *et al*. (2023) assert that GTP "produces outputs that are essentially indistinguishable from (even better than) what humans could produce". Additionally, ChatGPT can take account of the context and write a message according to the style of a well-known author (e.g., Shakespeare).

---

[1] On Dec. 27th 2023, *New York Times* sues OpenAI and Microsoft for copyright infringement on millions of its articles.

Given these assertions, this study aims to verify whether ChatGPT could generate political speeches for the US president (Hart, 2023). When asking ChatGPT to write a remark according to Reagan's style, can we still discriminate between the generated speech and the real one? If yes, what are the stylistic features that differ between the two versions? Can ChatGPT adopt a distinct shape when generating a message according to different US presidents or does it stay as a unique style? To answer those questions, this study will analyze their stylistic characteristics and compare them to four recent US presidents (namely Reagan, Clinton, Bush, and Obama) based on their *State of the Union* addresses.

In the remaining part of this article, we will first present some related work while Section 3 exposes the corpus used and presents some overall features. Section 4 analyzes some GPT stylistic features by comparing them to those occurring in speeches written by US presidents. Section 5 evaluates the global similarity between recent presidents and GTP. Section 6 provides some additional experiments based on characteristic vocabulary and Section 7 focuses on rhetorical and topical analysis. Finally, a conclusion reports the main findings of this study.

## 2 State of the Art

The domain and applications of stylometry cover a relatively large field from authorship attribution and profiling (Savoy, 2020), to the detection of plagiarism and fake documents as, for example, in criminology (Olsson, 2018), or even the dating of a document (Kreuz, 2023). In this context, the main objective of this study is to analyze and compare the style and rhetoric of political speeches automatically generated by ChatGPT (publicly available since Nov. 2022). The latter is mainly the user interface for dialog with the user while GPT is the underlying engine employed to generate the corresponding reports[2].

The LLM technology is grounded on a deep learning architecture (Goodfellow *et al*., 2016) based on a sequence of transformers with an attention mechanism (Vaswami *et al.* 2017). The most important notion to understand LLM is the following: Given a short sequence of tokens (e.g., words or punctuation symbols), the computer is able to automatically provide the next one. More precisely, knowing four tokens, the model must first determine the list of possible next tokens to complete the given sequence (Wolfram, 2023). For example, after the chain "the president of the", the computer, based on the training documents, can define a list of the next occurring token as {United, Philippines, Senate, US, firm, UK, republic, Ukraine, …}.

---

[2] In this study, the term GPT is used to denote the system but we have used GPT-3.5-turbo to generate our corpus.

From this list, and depending on some parameters[3], the system could then select the most probable one (e.g., "United" in our case), or based on a uniform distribution, one over the top *k* ranked tokens (e.g., "Senate") or, randomly depending on their respective probabilities of occurrence in the training texts (e.g., "US"). This non-deterministic process guarantees that the same request would produce distinct messages. All LLMs have been designed to generate plausible or reasonable sentences without checking the truth of the written facts. Thus as for all LLMs, GPT may include hallucinations in its answers (namely incorrect information, e.g., in our previous example, the sequence "the president of the UK" must be replaced by "the Prime Minister of the UK"). Moreover, the specification of the sources exploited to produce the text stays unknown to the user.

As mentioned previously, the main target application of such LLMs is to generate a short report to users' requests. To analyze this question, different studies expose the effectiveness of several learning strategies able to discriminate between answers generated by GPT-3 or written by human beings (Guo *et al.*, 2023). Based on a classifier trained on a given domain (e.g., RoBERTa), the recognition rate is rather high (around 95% to 98%). Such effectiveness is even obtained when the target language is not English (e.g., French in (Antoun *et al.*, 2023)). Such a high degree could be reduced when faced with a new and unknown domain or when substituting tokens by misspelled words (in such cases, the achieved accuracy rate varies from 28% to 60%). Of course, the message must include at least 1,000 letters to allow the detection system to reach such a small error rate.

With a similar objective, the CLEF-PAN 2019 international evaluation campaign evaluated different systems to automatically detect whether a set of tweets was generated by bots or by humans (Daelemans *et al.*, 2019). In this case too, the effectiveness was rather high (between 93% to 95% for the best approaches). However, the tweets written by bots where not produced by a LLM but corresponded to messages either containing a well-known citation, a passage of the Bible, or text corresponding to a predefined pattern (e.g., list of positions available in a large company).

## 3 Corpus Overview

To compare the written style of recent US presidents with speeches generated by machine, we asked ChatGPT to generate the *State of the Union* (SOTU) address for four presidents, namely Reagan, Clinton, Bush, and Obama. For each US leader, only the SOTU addresses have been taken into consideration. As the proper style of an author cannot be precisely defined *per se*, this study compares similar texts (SOTU) having the same genre, written in a same time period, and discussing similar topics for the same audience.

To help the freely available version 3.5 of GPT in its generative process, we provide for each president some written examples but not from any SOTU. Those instances have been extracted from the Miller website (`millercenter.org/president/`) containing a selection of the most famous presidential speeches[4]. All those SOTU messages have been downloaded from the website `www.presidency.ucsb.edu`.

---

[3] In this study, the temperature was fixed at 0.5, top_p at 0.4, frequency and presence penalty at 0.

To obtain a general overview of this political corpus, Table 1 shows some statistics about the speeches generated by GPT together with six US presidents (from R. Reagan to J. Biden)[5]. In Table 1, the number of speeches is given under the column with the label "Number". As word-tokens, we count the number of words, numbers and punctuation symbols. The final column (labelled "Types") specifies the number of distinct word-types included in those speeches. Based on GPT, we generated the SOTU addresses only for four presidents, namely Reagan, Clinton, Bush, and Obama. For each selected president, the generated allocations appear under the label "Reagan-GPT", "Clinton-GPT", etc.

**Table 1.** Some statistics about our American corpus

|   | Presidency | Number | Tokens | Types |
|---|---|---|---|---|
| Reagan-GPT |  | 7 | 9,255 | 1,449 |
| Clinton-GPT |  | 8 | 8,607 | 1,268 |
| Bush-GPT |  | 8 | 8,768 | 1,387 |
| Obama-GPT |  | 8 | 10,959 | 1,611 |
| R. Reagan | 1981–1989 | 7 | 37,004 | 3,514 |
| B. Clinton | 1993–2000 | 8 | 67,445 | 4,000 |
| W.G. Bush | 2001–2008 | 8 | 45,818 | 3,641 |
| B. Obama | 2009–2016 | 8 | 61,034 | 4,013 |
| D. Trump | 2017–2020 | 4 | 25,776 | 3,323 |
| J. Biden | 2021–2024 | 3 | 20,418 | 2,461 |

As one can see, the length of speeches returned by GPT are shorter than the real ones. One must recall that this online service is free and the generation of text requires intensive computer resources. Moreover, the used prompt does not provide a long list of information about the possible content of the target speech. For example, we wrote "Based on the given example [*a speech uttered by Reagan in 1984*], can you write a speech corresponding to the *State of the Union* in 1985 under Reagan's administration and with Reagan's style".

According to (Biber & Conrad, 2009), a stylistic study should be based on ubiquitous and frequent words. Following this principle, Table 2 shows the top ten most frequent lemmas used by GPT and the six US presidents. As one can see, both sets are identical. Clearly GPT has the ability to achieve some similarities with the presidential speeches. The ranking differs, with a higher frequency for the lemma "we" {we, us, our, ours} in machine-generated texts vs. true ones (6% vs. 3.8%). This is a significant aspect because the lemma "we" is frequently employed and common to all presidents (or prime ministers) in power. This pronoun has the advantage of being ambiguous; we are never sure who is behind the "we". Is it the president and his cabinet, or with the Congress or, more generally, the president and the people listening to the speech? In this last case, the speaker wants to establish a relationship with the audience, usually to imply it in the proposed solution.

---

[4] Of course, we don't know precisely the training sample employed by GPT and one might assume that many presidential speeches have been included. However, those messages, if appearing in the training set, are used to define the occurrence probability of a token given the four previous ones and not to identify a presidential style.

[5] As texts appearing in the training set have been published before 2019, we don't asked GPT to write SOTU addresses for Trump and Biden.

**Table 2a.** The top ten most frequent lemma on the American corpus subdivided by generated and real speeches

|    | GPT |       | US Presidents |        |
|----|-----|-------|---------------|--------|
| 1  | we  | 6.0%  | ,             | 5.0%*  |
| 2  | ,   | 5.7%  | .             | 4.5%*  |
| 3  | the | 5.7%  | the           | 4.2%*  |
| 4  | .   | 4.2%  | we            | 3.8%*  |
| 5  | of  | 3.5%  | and           | 3.3%   |
| 6  | and | 3.3%  | to            | 3.2%   |
| 7  | to  | 3.1%  | be            | 2.7%*  |
| 8  | an  | 2.5%  | of            | 2.4%*  |
| 9  | be  | 2.3%  | an            | 1.9%*  |
| 10 | in  | 1.7%  | in            | 1.6%   |

Moreover, GPT also tends to use more frequently the definite article "the" than the true presidents. Table 2 shows a percentage of 5.7% under GPT's pen vs. 4.2% employed by the ghostwriters at the White House. The preposition "of" follows the same pattern, a first indication that GPT favors noun phrases over the verbal ones.

To determine statistically whether or not a given percentage could be different than that produced by GPT, the proportion test (Conover, 1990) has been applied with the null hypothesis $H_0$ specifying that both population proportions are equal (bilateral test). For example, in Table 2 the proportion of "the" under GPT's pen is 5.7% vs. 4.2% with the presidents. Can we assume that both population proportions are equal? To verify this, the significance level $\alpha$ has been fixed at 1% and an asterisk '*' denotes an observation for which the null hypothesis must be rejected.

As one can in Table 2, a star has been added after "the" (under the label "US Presidents") indicating that the difference in proportions must be viewed as statistically significant. The $H_0$ assumption must also be rejected for other lemmas such as "we", "of", "be", "an" and the two punctuation symbols. The same statistical procedure has been applied in the following tables.

Various studies have analyzed the relative frequencies of personal pronouns revealing different stylistic and psychological traits about the author (Pennebaker, 2011). To investigate this facet, Table 3 shows the relative frequency of categories corresponding to personal pronouns. For example, under the category *Self*[6], one can find the word-types {I, me, mine, my, myself}. In this table, the smallest values are depicted in italics and the largest in bold.

---

[6] In this study, the denomination of a wordlist is presented in italics.

**Table 3.** Frequency of categories corresponding to personal pronouns

| Category | GPT | Reagan | Clinton | Bush | Obama | Trump | Biden |
|---|---|---|---|---|---|---|---|
| *Self* | *0.78%* | 0.92% | 1.38%* | 0.85% | 1.16%* | 1.00%* | **2.03%*** |
| *You* | *0.35%* | 0.44% | 0.69%* | 0.57%* | 0.48%* | 0.70%* | **1.17%*** |
| *She* | *0.00%* | 0.09%* | 0.10%* | 0.06%* | 0.15%* | **0.19%** | 0.19%* |
| *He* | *0.01%* | 0.11%* | 0.19%* | 0.20%* | 0.22%* | **0.59%** | 0.32%* |
| *We* | **6.02%** | 3.76%* | 3.97%* | 3.65%* | 3.78%* | 3.16%* | *2.46%** |
| *They* | *0.47%* | 0.75%* | **1.17%*** | 1.05%* | 0.99%* | 0.80%* | 1.04%* |

Table 3 indicates that GPT avoids using personal pronouns except the "we" which is overused. As depicted in Table 3, those proportions obtained by GPT are usually shown in italics. One can explain those low rates by the difficulty of being certain in establishing the correct pronominal anaphora (the link between the referent and the corresponding pronoun). With Biden, the lemma "we" is employed less frequently, but a higher intensity occurs with the category *Self* and *You*. This choice denotes the willingness to establish a relationship between the speaker and the audience. This aspect renders Biden's style distinct from that adopted by GPT. Moreover, the proportion differences are usually statistically significant as indicating by an asterisk.

## 4 Stylometric Analysis

According to previous studies, the written style could be characterized by certain numbers such as the mean word length with higher the mean, higher complexity. As a second measurement, one can take account of the vocabulary richness measured by the type-token ratio (TTR) (Hart, 1984). As other values, one can consider the hapax proportion (percentage of words occurring once), the lexical density (Biber *et al.*, 2002) or the percentage of big words (BW) defined as words composed of six letters or more (at least for the English language). For example, one can observe that some terms are easier to understand than others as, for example, between "ads" and "advertisings" or "desks" and "furniture". Such a relationship between complexity and word length is clearly established:

> "One finding of cognitive science is that words have the most powerful effect on our minds when they are simple. The technical term is basic level. Basic-level words tend to be short. ... Basic-level words are easily remembered; those messages will be best recalled that use basic-level language." (Lakoff & Wehling, 2012, p. 41)

For example, L. B. Johnson (presidency: 1963–1969) recognized the fear of having a too complex style by specifying to his ghostwriters: "I want four-letter words, and I want four sentences to the paragraph." (Sherrill, 1967).

Finally, to reflect the stylistic aspect related to the syntax, one can consider the mean sentence length (MSL), with a higher mean signaling a more detailed argumentation and usually less easy to follow.

To compare the real speeches with their GPT counterparts, Table 4 displays four general stylistic measurements starting with the average word length (column labelled "Word length")[7]. In the remaining columns, we have assessed the percentage of big words (BW), the TTR (or more precisely the moving average TTR denoted MATTR), and the mean sentence length (MSL). In the last two rows, the average over the six groups of presidential addresses and four GPT versions are displayed. For each measurement, the maximal value is depicted in bold and the smallest in italics.

**Table 4.** Some overall stylistic measurements of our corpus

|  | Word length | BW | MATTR | MSL |
|---|---|---|---|---|
| Reagan | 4.55* | 29.3%* | 0.347* | 21.04* |
| Reagan-GPT | 4.89 | 37.6% | *0.301* | 22.95 |
| Clinton | 4.34* | 26.8%* | 0.303 | 21.33* |
| Clinton-GPT | **4.96** | **38.4%** | 0.306 | **23.74** |
| Bush | 4.50* | 30.1%* | 0.329* | 20.08 |
| Bush-GPT | 4.83 | 36.7% | *0.301* | 21.06 |
| Obama | *4.31*  | *25.9%* | 0.323 | 19.76* |
| Obama-GPT | 4.91 | 36.9% | 0.315 | 22.74 |
| Mean president | 4.39* | 27.7%* | **0.359*** | *19.71*  |
| Mean GPT | 4.90 | 37.4% | 0.311 | 22.62 |

To reflect the language complexity, Hart (1984) suggests computing the mean word length. According to this measurement and for each president, the GPT version presents a significant higher mean suggesting that the messages tend to be more complex. In average, GPT's words contain 4.9 letters while the presidential words include, in average, 4.39 characters.

To confirm this first finding, one can also analyze the percentage of BWs depicted in the third column. In this case, the differences between the real speeches and the GPT output are relatively large and statistically significant. For example, when Clinton utters, in mean, 26.8% of long words, the GPT version contains 38.4% (a significant increase of 43.7%). The overall mean presented in the last two rows confirms this significant disparity.

In the fourth column, the type-token ratio (TTR) (Baayen, 2008) is provided. A high value specifies the presence of a rich vocabulary showing that the underlying text covers many different topics or that the author presents a theme from several points of view with different formulations. To compute this value, one divides the vocabulary size (number of word-types) by the text length (number of tokens). This estimator has the drawback of being unstable, tending to decrease with text length (Baayen, 2008). To avoid this problem, the computation could be based on a moving average (Covington & McFall, 2010). In this study, those values have been computed as an average value per 2,000 non-overlapping tokens. Thus, the first segment begins with the first word to the word appearing at position 2,000 while the second chunk starts at position 2,001 to 4,000, etc. This strategy is applied to avoid comparing texts with different lengths, a feature rendering the direct comparison between various measurements difficult.

---

[7] For this table, the proportion test has been replaced by the *t*-test with the same significance level of 1%.

Under this measurement, GPT tends to show a significant lower value (0.311 vs. 0.359), an indicator of the presence of a poorer vocabulary. However, for one president (Clinton), one can see the opposite, with the presidential addresses depicting a lower lexical level (a difference that cannot be viewed as statistically significant).

The last column of Table 4 shows the mean sentence length (MSL). In this case too, the GPT version is more complex than a presidential message. For Obama, for example, GPT's sentences contain, in mean, 22.74 words vs. 19.76 (a statistically significant difference). Producing longer sentences increases the language complexity (Hart, 1984) and renders the message harder to be understood by the audience.

To discriminate between the style of several authors, the distribution of the different part-of-speech (POS) categories could provide a new light on author's idiosyncrasies. Table 5 displays these distributions for six presidents and a mean for all speeches generated by GPT. In this table, the largest values are depicted in bold and the smallest in italics. We ignore some punctuation symbols (e.g., : " $) as well as numbers, foreign words, interjections (e.g. "yes", "euh"), possessive ending ('s), and the "to" before the infinitive. Thus, the total of each column does not reach the 100% value.

**Table 5.** Part-of-speech distribution over the US presidents and GPT

| POS | GPT | Reagan | Clinton | Bush | Obama | Trump | Biden |
|---|---|---|---|---|---|---|---|
| . | *3.95%* | 5.30%* | 4.48%* | 4.74%* | 4.82%* | 5.28%* | **5.91%*** |
| , | **6.86%** | 5.42%* | 4.93%* | 4.99%* | 4.99%* | 5.79%* | *4.88%*** |
| Conjunction | **5.43%** | 4.45%* | 3.72%* | 4.66%* | 3.95%* | 3.81%* | *3.60%*** |
| Article | 8.43% | **9.48%*** | 8.72% | 8.54% | 8.58% | 7.68%* | *7.65%*** |
| Preposition | 9.88% | 9.92% | 10.04% | **10.19%** | 10.06% | 9.46% | *9.10%*** |
| Pronoun | *7.42%* | 8.24%* | **8.99%*** | 7.34% | 8.46%* | 7.95% | 8.93%* |
| Adjective | **9.66%** | 6.42%* | 6.87%* | 6.95%* | 6.53%* | 7.39%* | *6.07%*** |
| Noun | **23.52%** | 19.04%* | 19.23%* | 20.53%* | 19.11%* | 17.96%* | *17.75%*** |
| Name | *2.78%* | 3.70%* | 3.92%* | 5.07%* | 3.54%* | **6.54%*** | 4.58%* |
| Adverb | *2.67%* | 4.44%* | 4.56%* | 3.52%* | 5.19%* | 5.20%* | **5.21%*** |
| Modal | 1.91% | **2.19%** | 2.18%* | 1.98% | 1.87% | 1.39%* | *1.25%*** |
| Verb | *13.46%* | 14.93%* | 15.30%* | 15.62%* | 16.44%* | 14.98%* | **16.67%*** |

This table shows that GPT uses less periods (3.95%), a clear indication that the mean sentence length is larger than for the US presidents (see also Table 4). On the other hand, the comma is overused (6.86%), usually in the context of a sequence of nouns. As indicated in Table 5, conjunction, nouns, and adjectives appear with a higher frequency under GPT's pen. This distribution of three POS categories corresponds to a descriptive style favoring noun phrases with a language oriented towards a narrative of the situation or providing a general explanation.

On the other hand, Biden or Obama presents the highest frequency of verbs (16.67% and 16.44%) and adverbs, with a resulting tone towards action. When considering names, adverbs, modal verbs or verbs in general, these POS categories tend to be employ less frequently by GPT.

In Table 5, the stylistic difference between presidents can also be detected. Both Biden and Trump are writing with short sentences (full stop occurring with respectively 5.91% and 5.28%). Biden employs verbs, adverbs, and pronouns frequently with a higher frequency of verb phrases favoring an active message. On the opposite end, one can find Reagan favoring articles, and nouns. A second characteristic of Trump's style is the presence of many commas (5.79%) and names (6.54%). This high intensity of names corresponds to the recurrent use of "America", "American", "United States" or "Congress".

## 5 Characteristic Vocabulary

As all presidents are speaking with similar terms, the differentiation between them resides in their frequencies. To detect those differences, one can take account of the characteristic vocabulary belonging to each author. To determine the terms overused (or under-used) by a writer, Muller (1992) suggests analyzing the number of term occurrences between a specific author compared to the whole corpus.

Mainly, the underlying idea is the following. For a given term $t_i$, we compute its occurrence frequency both in the set $P_0$ (value denoted $tf_{i,0}$) and in the second part $P_1$ (denoted $tf_{i,1}$). The set $P_0$ would be the sample written by the target author, while $P_1$ the rest of the corpus. Thus, for the entire corpus the occurrence frequency of the term $t_i$ becomes $tf_{i,0} + tf_{i,1}$. The total number of word tokens in part $P_0$ (or its size) is denoted $n_0$ similarly with $P_1$ and $n_1$. Thus, the size of the entire corpus is defined by $n = n_0 + n_1$. For any given term $t_i$ the distribution is assumed to be binomial, with parameters $n_0$ and $p(t_i)$ representing the probability of the term $t_i$ being randomly selected from the entire corpus. Based on the maximum likelihood principle, this probability would be estimated as follows:

$$p(t_i) = \frac{tf_{i,0} + tf_{i,1}}{n} \quad (1)$$

Through repeating this drawing $n_0$ times we are able to estimate the expected number of occurrences of term $t_i$ in part $P_0$ using the expression $n_0 \cdot p(t_i)$. We can then compare this expected number to the observed number (namely $tf_{i,0}$), where any large differences between these two values indicate a deviation from the expected behaviour. To obtain a more precise definition of large we account for variance in the underlying binomial process (defined as $n_0 \cdot p(t_i) \cdot (1-p(t_i))$). Equation 2 defines the final standardized Z score (or standard normal distribution $N(0,1)$) for term $t_i$, using the partition $P_0$ and $P_1$:

$$\text{Z score}(t_{i,0}) = \frac{tf_{i,0} - n_0 \cdot p(t_i)}{\sqrt{n_0 \cdot p(t_i) \cdot (1-p(t_i))}} \quad (2)$$

For each selected term, we apply this procedure to weight its specificity according to the underlying text excerpt $P_0$. Based on the Z score value, we then verify whether this term is used proportionally with roughly the same frequency in both parts (Z score value close to 0). On the other hand, when a term is assigned a positive Z score larger than a given threshold (e.g., 3), we consider it over-used or belonging to the specific vocabulary of $P_0$. The result of this procedure appears in Table 6 in which the top ten most characteristic words used by GPT and four US presidents.

When focusing on each presidency, a closer look reveals some of the recurrent problems faced by each corresponding administration. The Soviet Union, nuclear weapon reduction, the promotion of freedom, and the spending for defense (Star War) with Reagan while Clinton's two terms can be characterized by welfare, education for children and the fight against crime. With Bush, one can recall the war against Iraq (Saddam Hussein) and the terrorists (Al Qaida) while with Obama, the focus is on jobs (the 2008 crisis) and energy.

Table 6. Terms overused by GPT and the US presidents

| Rank | Reagan | | Clinton | | Bush | | Obama | |
|---|---|---|---|---|---|---|---|---|
| | GPT | real | GPT | real | GPT | real | GPT | real |
| 1 | principle | soviet | shared | welfare | adversity | terrorist | healthcare | that |
| 2 | pursuit | freedom | fostering | child | we | Iraq | investing | business |
| 3 | commitment | govern. | economic | work | unwavering | Iraqi | rural | why |
| 4 | dedication | program | collaboration | ought | remains | enemy | ensuring | job |
| 5 | utmost | defense | fiscal | should | commitment | terror | journey | energy |
| 6 | nation | spending | commitment | all | terrorism | Qaida | affordable | or |
| 7 | unwavering | reduction | business | crime | resolve | Iraqis | gender | how |
| 8 | enduring | arm | where | must | resilience | Saddam | particularly | kid |
| 9 | navigate | needy | boundless | to | Iraqi | Al | broadband | college |
| 10 | strength | federal | understanding | parent | waver | Hussein | challenge | like |

Comparing the GPT version for each presidency with the true speeches, one can see that the topics are not quite similar. Looking at the most characteristic words under GPT's pen, one can see that the adjectives are rather general (shared, fiscal, rural, economic) instead of centering on more specific problems. The nouns follow the same general perspective (commitment, challenge, resilience) with a recurrent neutral tone avoiding divisive formulations (e.g., collaboration). To emphasize this feature, one cannot observe a clear time and space anchoring. Very few city, country or proper names appear in the characteristic vocabulary employed by GPT. Therefore, its messages appear out-of-time and space, reinforcing the impression of a general and neutral tone. One can however mention some relationship between problems faced by a particular president. For example, one can see "terrorism" or "Iraqi" employed by GPT following Bush's style or "healthcare" under Obama.

Knowing the terms overused by an author, his/her typical sentences can be defined as those containing the largest number of overused words. As an example, GPT generating the *State of the Union* for Bush in 2002 produces the following passage in which the over-used terms appeared in bold:

"On **September 11**th, **2001**, **our great** land **faced** an unprecedented and unprovoked attack on **our freedom**, **our** values, and **our** way **of** life". (GPT according to Bush's style, 2002)

With this example, one can see that GPT could generate a well-adapted sentence reflecting a main concern of Bush's administration. Moreover, GPT clearly knows how to start a SOTU address with the appropriate salutations, or to close it with a conventional greeting:

"Ladies and gentlemen, members of Congress, distinguished guests, and my fellow Americans. Thank you for joining me tonight for this State of the Union address
…
May God bless you all, and may God continue to bless the United States of America."
(GPT according to Bush's style for 2001)

## 6  Rhetorical and Topical Analysis

As leaders tend to reuse the same words, we can hypothesize that some presidents will talk more on some topics, or using some rhetorical forms, more often than others. To analyze such regular usage of some expressions, one can regroup related semantical words under a given tag. For example, in this study, the category *Symbolism* (Hart *et al*., 2013) contains terms related to the country (e.g., nation, America), ideology (e.g., democracy, freedom, peace), or generally, political concepts and institutions (e.g., law, government). Under *Tenacity* (e.g., was, is, will, etc.), one can study the absolute confidence of the author's declaration or claim (Hart, 1984) while the *Blame* category includes terms such as angry, deceptive, incompetent, etc. In this study, words belonging to the *Achieve* tag (e.g., first, plan, win, …) form another dedicated category.

Grounded on such wordlists, Hart (1984) portrays the rhetorical and stylistic differences between the US presidents from Truman to Reagan. In a second study, Hart *et al*. (2013) exposed the stylistic changes from G.W. Bush to Obama as well as some features associated with Trump's presidency (Hart, 2020). Recently, Hart (2023) exposed the change in style and eloquence of US presidents and politicians during the last century.

As an alternative, one can adopt the LIWC (Linguistic Inquiry & Word Count) system (Tausczik & Pennebaker, 2010) organizing expressions under syntactical, emotional or psychological categories. These classes may match grammatical categories (e.g., first person singular denoted by *Self*, see Table 3), broader ones (e.g., verbs) as well as specific ones (verbs in the past tense, auxiliary verbs). On the semantics level, the LIWC system defines positive emotions (*Posemo*) (e.g., happy, hope, peace), negative ones (*Negemo*) (e.g., fear, blam*[8]), or terms related to *Humans* (e.g., family, woman, child*).

Based on these two approaches, Table 7 reports the relative percentage of seven wordlists describing different semantic and rhetorical facets. In this table, the largest values are shown in bold and the smallest appear in italics. These wordlists have been chosen to reflect the emotional content of speeches (e.g., *Posemo*, *Negemo*), references to political entities (*Symbolism*), as well as the use of certain tones (e.g., *Tenacity*), arrogance (*Blame*) or relations with humans.

---
[8] When generating an entry in a wordlist, one can use the symbol '*' to denote any sequence of letters.

As shown in Table 7, GPT regularly appears with the lowest or highest percentage in these categories. In the *Posemo* row, one can see that presidents prefer to opt for positive attitudes and in this view, Bush also presents a high percentage (3.78%, mainly with patterns 'secur*', 'freed*', and 'peace'), but lower than the highest achieved by GPT (4.67%) with many occurrences related to 'challeng*' and 'promis*'. On the other hand, GPT presents the lowest occurrence rate for negative expressions (0.95%), a rate lower than for all the presidents.

**Table 7.** Percentage of some selected wordlists

| Category | GPT | Reagan | Clinton | Bush | Obama | Trump | Biden |
|---|---|---|---|---|---|---|---|
| *Posemo* | **4.67**% | 3.47%* | 3.11%* | 3.78%* | *2.78%** | 3.38%* | 2.79%* |
| *Negemo* | *0.95%* | 1.14% | 0.97% | 1.50%* | 1.07%* | **1.53**%* | 1.14% |
| *Symbolism* | 4.40% | 3.84%* | 3.52%* | 4.10% | *3.16%** | **4.43**% | 3.50%* |
| *Tenacity* | *4.76%* | 6.26%* | 5.94%* | 6.21%* | 6.49%* | 6.46%* | **6.86**%* |
| *Blame* | *0.20%* | 0.44%* | 0.42%* | 0.40%* | 0.46%* | **0.60**%* | 0.39%* |
| *Humans* | 0.73% | 0.57%* | 0.80% | **0.94**%* | 0.55%* | 0.83% | *0.37%** |
| *Achieve* | **3.61**% | 2.56%* | 2.87%* | 2.51% | 2.39%* | 2.27%* | *1.96%** |

In politics, the *Symbolism* list contains general political terms, and Trump uses them more frequently (4.43%) than other presidents. GPT stays at the same high level though (4.4%), emphasizing the abstraction with references to political institutions or to the nation (e.g., America). This could be interpreted as producing a more abstract pitch.

The certainty tone (*Tenacity*) is also an important rhetorical attitude for a leader in power. The president must convince the audience that he knows how to solve problems. Biden's messages contain the highest level of certainty (6.86%)while GPT appears to be the lowest in this category (4.76%).

The terms belonging to the class *Blame* (terms indicating social inappropriateness) are also not employed by GPT (0.2%). In this view, Trump presents the highest score (0.6%) as well as for the negative emotional words (1.53%). When analyzing words related to humans, GPT has a relative frequency comparable to the presidents, a value close to those obtained by Clinton or Trump. Finally, the category *Achieve* (plan, win, first, etc.) can be used to verify that the president has clear plans and is achieving concrete results. In this case, GPT remains with the highest occurrence frequency (3.61%) while Biden shows the lowest (1.96%).

Of course, one must recognize that automatically extracting tone, psychological traits or topics from speeches is not exempt from difficulties. For example, the word "joy" could be associated to a positive emotion, as well as to indicate the title of a movie (produced in 2015), a clothing company, an abbreviation, or a website. It is known that all natural languages have numerous ambiguities, rendering the automatic extraction of sentiments or topics subject to caution.

However, such general ambiguity could exist to a lesser extent in political speeches than in a more general context (e.g., small talk).

## 7  Intertextual Distance

To evaluate the similarity between a pair of texts, Labbé (2007) proposed an intertextual distance based on the entire vocabulary. The computation of this measure between Text A and Text B is defined according to Equation 3 where $n_A$ indicates the length of Text A (in number of words), and $tf_{i,A}$ denotes the absolute frequency of the *i*th term (for $i = 1, 2, …, m$). The value *m* represents the vocabulary length. Usually both texts do not have the same length, so let us assume that Text B is the longer. To reduce the longer text to the size of the smaller, each of the term frequencies (in our case $tf_{i,B}$) is multiplied by the ratio of the two text lengths, as indicated in the second part of Equation 3.

$$D(A, B) = \sum_{i=1}^{m} \left| tf_{i,A} - \widehat{tf_{i,B}} \right| \Big/ (2 \cdot n_A) \qquad \text{with } \widehat{tf_{i,B}} = tf_{i,B} \cdot n_A / n_B \qquad (3)$$

This formulation returns a value between 0 and 1 depending on the lexical overlap between two texts. When two documents are identical, the distance is 0. The largest distance of 1 would appear when the two speeches have nothing in common (e.g., one written in Chinese and the other in English). Between these two limits, the distance value depends on the number of terms appearing in both speeches, and their occurrence frequencies. Even if this computation can take account for different text lengths, it is reasonable to limit the difference to eight times the smallest.

Based on this measure, one can compute the intertextual distance between the six US presidents and messages written by GPT. For four presidents, we subdivide their speeches according to their two terms in office. For example, Obama's speeches are subdivided into "ObamaA" (his first term in the White House) and "ObamaB" (last four years). In addition, the GPT speeches have also been split according to the two terms. Table A.1 depicts the resulting text lengths for these 18 subdivisions.

Directly displaying the 18 x 18 matrix containing these distances has a limited interest. Knowing that this matrix is symmetric and that the distance to itself is nil, we still have ((18 x 18) – 18) / 2 = 153 values. To achieve a better picture than a dendrogram, such distance matrices can be represented by a tree-based visualization respecting *approximately* the real distances between all nodes (Bartélémy & Guénoche, 1991), (Baayen, 2008), (Paradis, 2011). We adopt this new representation and the result is displayed in Figure 1 (produced by the R software).

In this figure, the distance between two presidents is indicated by the lengths of the lines connecting them. For example, starting with ObamaB, one can follow the branch until reaching the intersection, then one can go along the lines leading to the second person (e.g., BushA or ReaganB). In this figure, the shortest distance occurs between ObamaA and ObamaB (0.179), and other short distances can also be found with the other presidents elected for two terms (e.g., ClintonA–ClintonB: 0.197; BushA–BushB: 0.191, ReaganA–ReaganB: 0.21). The longest distance links Biden with ReaganB GPT (0.437) or with Biden—ClintonA GPT (0.436) or Biden–BushA GPT (0.425).

**Figure 1.** Intertextual distance in our American corpus

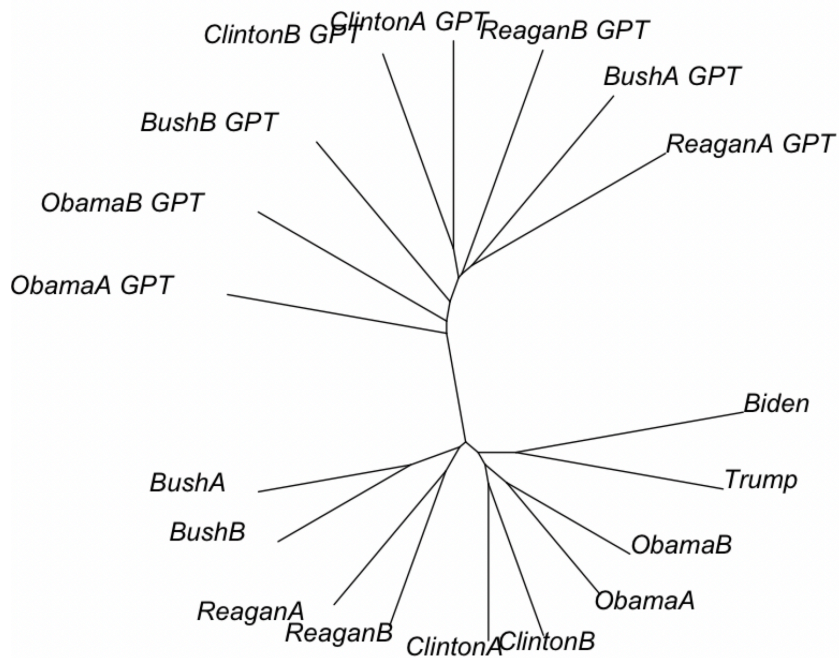

Overall, this figure illustrates the large overall difference between GPT speeches on the one hand (on the top part) and the six recent US presidents on the other (in the bottom). The speeches generated by GPT clearly present a distinctive style, and no clear relationship can be established with the six most recent US presidents. When asking GPT to write a text according to the style of a given president (e.g., Obama), one can observe that the generated texts are closer to other generated addresses than to the target author (e.g., ObamaA–ObamaA GPT: 0.335, ObamaB–ObamaB GPT: 0.333). One must recall that the training set never includes a SOTU address but contains another presidential speech.

Moreover, one can see that the speeches related to the two terms are closer than the others, with one exception, BushA GPT appearing between ReaganA GPT and ReaganB GPT. Is this an error? Bush spent his first year at the White House before the terrorism attack of September 11[th]. In fact, before this date, Bush's speeches presented a clear relationship with Reagan's topics with a focus on economic questions, tax reduction and defense. The training speech provided to GPT for this first term reflects those topics. After September 11th, the Bush administration focused more on the terrorist questions, homeland security and war against Iraq. Therefore, Figure 1 mirrors this difference between Bush's two terms.

# 8 Conclusion

The various experiments performed in this study demonstrate that GPT can generate political speeches sharing similarities with real *State of the Union* addresses. For example, based on Table 2, the top ten most frequent words are the same between GPT versions and true speeches. Moreover, this generative system recognizes the importance of the pronoun "we" (see Table 2) also occurring with a high frequency under many leaders in power. This stylistic facet could also be viewed as a clear sign of possible close similarity with presidents (or prime ministers).

However, when analyzing the mean sentence length (MSL), GPT opts for longer constructions (MSL: 22.62) than the US presidents (e.g., 19.71) (see Table 4). GPT's language complexity is therefore higher and this aspect is increased by considering both the presence of longer words (mean word length of 4.9 vs. 4.39 for the mean presidential addresses) and a higher percentage of big words (BW: 37.4% compared to 27.7%).

To provide a better explanation, Table 5 highlights the POS distribution difference between texts written by a computer and those authored by humans. One can observe that GPT favors noun-phrases (nouns, articles, prepositions) and does not anchor its argumentation in space, time or with many names. In addition, Table 7 indicates that GPT's tone is clearly positive avoiding negative expressions. Moreover, GPT includes many references to political institutions and symbols (mainly with the term "nation" or "America").

As a result, the overall GPT writing style can be characterized by a didactic and neutral tone. One can add that the global impression is that of reading a report without real pitch. Many personal pronouns are missing (you, s/he, they) (see Table 3) and the presentation inclines to stay at a descriptive level, without taking any divisive position (see Table 6). Moreover, GPT avoids specific examples and never employs an argument that could cause clear disagreements between people.

Using an intertextual distance measure computed according to the whole vocabulary, a overall picture indicates that GPT addresses differ from the real ones (see Figure 1). True ghostwriters employed more distinct stylistic and rhetorical *modus operandi* than GPT. On the other hand, when asking GPT to write according to the style of a given president, the computer is able to produce allocutions that differ from one leader to the next (see Figure 1).

Finally, we need to mention that GPT is not the sole LLM system or a final product. Such models will evolve over time. Thus, with improvements allowing the system to include more personal pronouns of the second or third person, together with the inclusion of more names, or geographical anchors, GPT will generate speeches closer to real presidential addresses.

## References

Antoun, W., Mouilleron, V., Sagot, B., and Seddah, D. (2023). Towards a robust detection of language model-generated text: Is ChatGPT that easy to detect? *CORIA-TALN Conference*, Paris June 2023, 1–14.

Baayen, H.R. (2008). *Analyzing Linguistic Data. A Practical Introduction Using R*. Cambridge: Cambridge University Press.

Bartélémy, J.P., and Guénoche, A. (1991). *Trees and Proximity Representations*. New York: John Wiley

Biber, D., Conrad, S., and Leech, G. (2002). *The Longman Student Grammar of Spoken and Written English*. London: Longman.

Biber, D., and Conrad, S. (2009). *Register, Genre, and Style*. Cambridge: Cambridge University Press.

Bubeck, S., Chandrasekaran, V., Eldan, R., Gehrke, J., Horvitz, E., Kamar, E., Lee, P., Lee, Y.T., Lundberg, S., Nori, H., Palangi, H., Ribeiro, M.T., and Zhang, Y. (2023). Sparks of artificial general intelligence: Early experiments with GPT-4. arXiv:2303.12712v5.

Conover, W.J. 1990. *Practical Nonparametric Statistics*. New York: John Wiley & Sons.

Covington, M.A., and McFall, J.D. 2010. Cutting the Goridan knot: The moving-average type-token ratio (MATTR). *Journal of Quantitative Linguistics*, 17(2):94-100.

Daelemans, W., Kestemont, M., Manjavacas, E., Potthast, M., Rangel, F., Rosso, P., Specht, G., Stamatatos, E., Stein, B., Tschuggnall, M., Wiegmann, M, and Zangerle, E. (2019). Overview of PAN 2019: Bots and Gender Profiling, Celebrity Profiling, Cross-Domain Authorship Attribution and Style Change Detection. In Experimental IR Meets Multilinguality, Multimodality, F. Crestani, M. Braschler, J. Savoy, A. Rauber, H. Müller, D. Losada, G. Heinatz Bürki, L. Cappelo, and N. Ferro (Eds), Springer, Cham, Lecture Notes in Computer Science #11696, 402–416.

Goodfellow, I., Bengio, Y., and Courville, A. (2016). *Deep Learning*. Boston: The MIT Press.

Guo, B., Zhang, X., Wang, Z., Jiang, M., Nie, J., Ding, Y., Yue, J., and Wu, Y. (2023). How close if ChatGPT to human experts? Comparison corpus, evaluation and detection. arXiv:2301.07597.

Hart, R.P. (1984). *Verbal Style and the Presidency. A Computer-based Analysis*. Orlando: Academic Press.

Hart, R.P., Childers, J.P., and Lind, C.J. (2013). *Political Tone. How Leaders Talk and Why*. Chicago: The University of Chicago Press.

Hart, R.P. (2020). *Trump and Us. What He Says and Why People Listen*. Cambridge: Cambridge University Press.

Hart, R.P. (2023). *American Eloquence: Language and Leadership in the Twentieth Century*. New York: Columbia University Press.

Jiao, W., Wang, W., Huang, J.T., Wang, X., and Tu, Z. (2023). Is ChatGPT a good translator? Yes with GPT-4 as the engine. arXiv: 2301.08745. https://arxiv.org/pdf/2301.08745.pdf

Karsdorp, F., Kestemont, M., and Riddell, A. (2021). *Humanities Data Analysis. Case Studies with Python*. Princeton: Princeton University Press.

Kreuz, R. (2023). *Linguistics Fingerprints. How Language Creates and Reveals Identity*. Prometheus Books, Guilford, (CT).

Labbé, D. (2007). Experiments on authorship attribution by intertextual distance in English. *Journal of Quantitative Linguistics*, 14(1), 33–80.

Lakoff, G., and Wehling, E. (2012). *The Little Blue Book: The Essential Guide to Thinking and Talking Democratic*. New York: Free Press.

Muller, C. (1992). *Principes et Méthodes de Statistique Lexicale*. Paris: Honoré Champion

Olsson, J. (2018). *More Wordcrime. Solving Crime Through Forensic Linguistics.* London: Bloomsbury.

Paradis, E. 2011. *Analysis of Phylogenetics and Evolution with R.* New York: Springer.

Pennebaker, J.W. (2011). *The Secret Life of Pronouns. What our Words Say About Us*. New York: Bloomsbury Press.

Savoy, J. (2020). *Machine Learning Methods for Stylometry. Authorship Attribution and Author Profiling*. Cham: Springer.

Sherrill, R. (1967). *The Accidental President*. New York: Grossman.

Tausczik, Y.R., and Pennebaker, J.W. (2010). The psychological meaning of words: LIWC and computerized text analysis methods. *Journal of Language and Social Psychology*, 29(1):24-54.

Vaswani, A., Shazeer, N., Parmar, N., Uszkoreit, J., Jones, L., Gomez, A.N., Kaiser, L., and Polosukhin, I. (2017). Attention is all you need. In Advances in Neural Information Processing Systems, 30.

Wolfram, S. (2023). *What is GPT-4 Doing… and What Does it Work?*. Orlando: Wolfram Research Inc., Champaign (IL).

Yule, G. (2020). *The Study of Language*. 7th ed., Cambridge: Cambridge University Press.

Zhao, W., Zhou, K., Li, J., Tang, T., Wang, X, Hou, Y. Min, Y., Zhang, B., Zhang, J., Dong, Z., Du, Y., Yang, C., Chen, Y., Chen, Z., Jiang, J., Ren, R., Li, Y., Tang, X., Peiyu, P., Nie, J.Y., and Wen, J.R. (2023). A survey of large language models. arXiv: 2303.18223v11.

# Annexe

**Table A.1.** Some statistics about our presidential corpus subdivided by term

|  | Presidency | Number | Tokens | Types |
|---|---|---|---|---|
| ReaganA-GPT |  | 3 | 4,167 | 935 |
| ReaganB-GPT |  | 4 | 5,088 | 982 |
| ClintonA-GPT |  | 4 | 3,923 | 860 |
| ClintonB-GPT |  | 4 | 4,684 | 877 |
| BushA-GPT |  | 4 | 4,627 | 914 |
| BushB-GPT |  | 4 | 4,141 | 893 |
| ObamaA-GPT |  | 4 | 6,049 | 1,172 |
| ObamaB-GPT |  | 4 | 4,910 | 997 |
| R. Reagan A | 1981–1984 | 3 | 18,061 | 2,365 |
| R. Reagan B | 1984–1989 | 4 | 18,943 | 2,515 |
| B. Clinton A | 1993–1997 | 4 | 33,990 | 2,700 |
| B. Clinton B | 1997–2000 | 4 | 33,455 | 2,931 |
| W.G. Bush A | 2001–2004 | 4 | 21,363 | 2,507 |
| W.G. Bush B | 2004–2008 | 4 | 24,455 | 2,681 |
| B. Obama A | 2009–2012 | 4 | 31,026 | 2,773 |
| B. Obama B | 2012–2016 | 4 | 30,008 | 2,958 |
| D. Trump | 2017–2020 | 4 | 25,776 | 3,323 |
| J. Biden | 2021–2024 | 3 | 20,418 | 2,461 |